\documentclass{vgtc}                          

\ifpdf
  \pdfoutput=1\relax                   
  \usepackage{graphicx}   
  \graphicspath{ {pictures/} }
  \DeclareGraphicsExtensions{.pdf,.png,.jpg,.jpeg} 
\else
  \usepackage{graphicx}                
  \DeclareGraphicsExtensions{.eps}     
\fi%

\graphicspath{{figures/}{pictures/}{images/}{./}} 

\usepackage{microtype}                 
\PassOptionsToPackage{warn}{textcomp}  
\usepackage{textcomp}                  
\usepackage{mathptmx}                  
\usepackage{times}                     
\usepackage{cite}                      
\usepackage{tabu}                      
\usepackage{booktabs}                  
\usepackage{multirow}
\usepackage[table,xcdraw]{xcolor}
\usepackage[font=small,labelfont=bf]{caption}

\usepackage{subfigure}

\onlineid{1053}

\vgtccategory{Research}

\vgtcinsertpkg

\title{Egocentric Gesture Recognition for Head-Mounted AR devices}

\author{Tejo Chalasani\thanks{e-mail: chalasat@scss.tcd.ie} %
\and Jan Ondrej\thanks{e-mail:jan@volograms.com} %
\and Aljosa Smolic\thanks{e-mail:smolica@scss.tcd.ie}}
\affiliation{\scriptsize V-SENSE, School of Computer Science and Statistics \\Trinity College Dublin}

\abstract{Natural interaction with virtual objects in AR/VR environments makes for a smooth user experience. Gestures are a natural extension from real world to augmented space to achieve these interactions. Finding discriminating spatio-temporal features relevant to gestures and hands in ego-view is the primary challenge for recognising egocentric gestures. In this work we propose a data driven end-to-end deep learning approach to address the problem of egocentric gesture recognition, which combines an ego-hand encoder network to find ego-hand features, and a recurrent neural network to discern temporally discriminating features. Since deep learning networks are data intensive, we propose a novel data augmentation technique using green screen capture to alleviate the problem of ground truth annotation. In addition we publish a dataset of 10 gestures performed in a natural fashion in front of a green screen for training and the same 10 gestures performed in different natural scenes without green screen for validation. We also present the results of our network's performance in comparison to the state-of-the-art using the AirGest dataset.%
} 

\CCScatlist{
  \CCScatTwelve{Egocentric Gesture Recognition}{Deep Learning}{LSTMs};
  \CCScatTwelve{Human Computer Interfaces}{Natural Gestures}{}{}
  }

\begin{document}

\firstsection{Introduction}
\maketitle

Interaction with virtual objects in Augmented Reality (AR) is a core principle for smooth user experience. Gestures represent a natural way of interaction and communication in our daily life. Thus recognising gestures and using them to interact with virtual elements is a natural extension from reality to AR. With commercially-viable wearable AR devices like Microsoft HoloLens, Magic Leap One, Daqri Smart Helmet being available to consumers the need for intuitive ways of interaction increases. The head-mounted AR devices have a camera that is placed between the eyes of the user giving an egocentric view of the world. Recognising egocentric gestures thus gives us a natural way to interface with the virtual elements displayed in such AR devices.

As different AR/VR applications may vary significantly regarding their user interactions, the set of gestures needed for smooth user interaction may also be very different. Therefore, solutions should also be easily adaptable and extensible for different applications. Having a large dataset with many different ego-hand gestures as defined in \cite{Zhang2018} is good to test different recognition algorithms. However, it does not address the challenge of easily adding or removing new gestures to the dataset, since a large number of users are required to perform the gestures in multiple different scenarios. Devices like HoloLens do support a small quantity of hand gestures, but they do not support addition of new gestures. In this paper not only do we address the issue of recognising ego-hand gestures, but also the ability to add new gestures easily to data driven deep learning networks with reduced amount of data.

\begin{figure}[tb]
 \centering 
 \includegraphics[width=0.8\columnwidth]{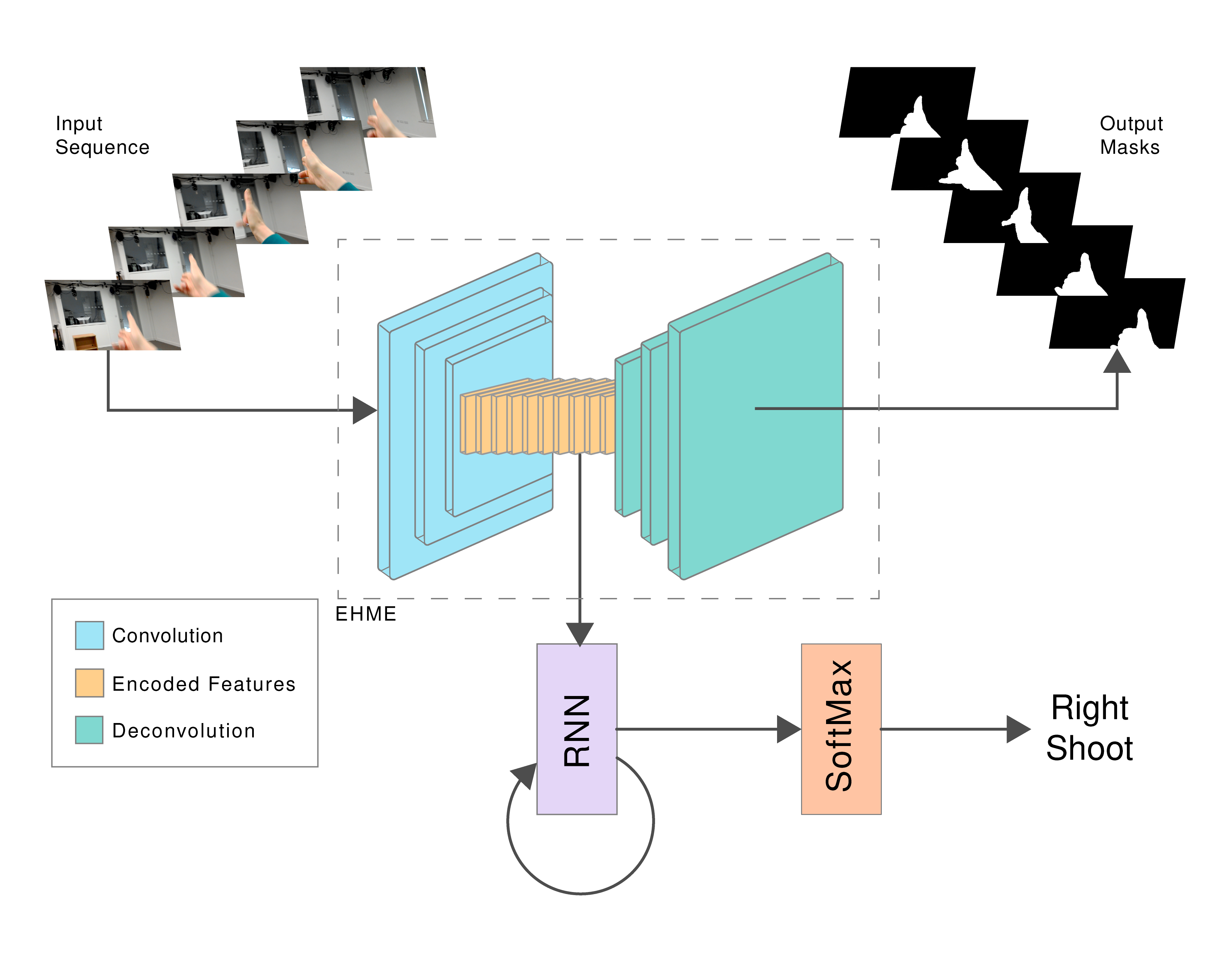}
 \caption{Network Architecture to recognise egocentric gestures: Ego-hand mask encoder (EHME) that encodes a sequence of input images and Recurrent Neural Network (RNN) to recognise the gesture from sequence of encoded images. }
 \label{fig:architecture}
\end{figure}

We propose an end-to-end learnable deep network architecture (Fig \ref{fig:architecture}) to encode ego-hand features and use them in a recurrent neural network (RNN) to find temporal features that can map a given arbitrary length image sequence to a gesture. Furthermore, we propose a new data augmentation process using a green screen masking technique as a way to create a large amount of data from a small set of captured data which is used to train our proposed network. In addition, as our approach is fully automatic, it also helps us to eliminate the extremely time-consuming and costly process of annotating each and every frame interactively for hand segmentation. Our proposed network not only simplifies the architecture, making it plausible to be used on future wearable devices, but also forgoes the usage of an optical flow branch and a 3D CNN, which is standard in many action and gesture recognition frameworks like \cite{Molchanov2016, Cao2017, Singh2016, Simonyan2014}.

Further, we make our dataset publicly available, which consists of gestures performed in front of a green screen by 22 different users recorded with an egocentric camera on HoloLens and the corresponding segmentation masks. The green screen dataset is augmented and used to train our network. Gestures performed in a normal environment (i.e. not in front of green screen, also provided in the dataset) are then used for testing. Each image in the database is also provided with egocentric camera 6DOF pose information.

Consequently, the main contributions of our work are the following:

\begin{itemize}
    \item We introduce a new deep network architecture that simultaneously encodes ego-hand features and finds temporal features for gesture recognition in RGB image sequences of an arbitrary length. The network is smaller compared to the existing architectures making it plausible to fit on future wearable AR devices.
    \item We introduce a new data augmentation technique using green screening to train from a small amount of captured data which also eliminates the need for manual ground truth hand mask annotation.
    \item We publish a green screen egocentric gesture dataset that can be used for training egocentric gesture recognition and is easily extendable.
\end{itemize}
It must be noted that the network architecture presented would be very slow to run on current mobile devices without specialised hardware to run deep neural networks, one of the aims of this work is to look for architectures smaller in comparison to existing ones but not necessary to run on existing mobile headsets.
\section{Related Work}

Vision-based gesture recognition traditionally used hand-crafted spatio-temporal features mainly in a rule based framework\cite{Cutler1998, Mu-ChunSu2000} or in a machine learning framework based on Hidden Markov Models (HMM) \cite{Moni2009}. Ego motion from the camera in AR devices adds another layer of complexity when compared to gesture recognition from static cameras. To deal with this additional complexity methods like \cite{Baraldi2014} suggest homography compensation as a preprocessing step before calculating hand crafted features like improved Dense Trajectories (iDT) \cite{Wang2013}.

Recent advances in Deep Learning techniques, general purpose GPU usage and the availability of large amounts of annotated data lead to tremendous improvements in various computer vision related problems like object detection \cite{Krizhevsky2012, Simonyan2014a, Wu2016}, tracking \cite{david2016, Posner2016} and object localisation \cite{Girshick2015}. To deal with the temporal component of activity and gesture recognition using deep learning there are two main approaches. The first one uses optical flow and a 3D CNN to extract spatio-temporal features and to pass them to an SVM or class entropy layer for classification \cite{Tran2015, Singh2016, Wang2017b, Simonyan2014, Molchanov2016}. The second uses the outputs of CNNs as inputs to RNN as these kind of networks are designed to inherently manage the temporal dimension\cite{Cao2017, Donahue2014}. 

Neurological research \cite{Goodale1992} suggests a two branch approach to action recognition, a ventral branch for object recognition, and a dorsal branch for motion recognition. This inspired the two stream approach for action recognition using deep learning \cite{Simonyan2014}, which was extended to egocentric action recognition by adding a third stream consisting of ego-hand segmentation information that was manually generated \cite{Singh2016}. Taking inspiration from the above work, instead of manually adding hand masks as input, we rather encode ego-hand masks automatically and use these encoded features in a second branch which consists of an RNN to deal with the challenge of recognising egocentric gestures from image sequences.

Zhang et al. \cite{Zhang2018} recently published a database of egocentric gestures specific to interactions in wearable devices. This dataset contains 83 different gestures performed by 50 users in various settings. However, we posit that gestures are continuously evolving and there is a need for adding new gestures. For a new gesture to be recognised in the framework proposed by \cite{Zhang2018, Cao2017} 50 different users have to perform the gesture in 8 different scenarios. This makes adding a new gesture for recognition a cumbersome task, which is one of the challenges we address in our work.

The method in \cite{Jain2017} uses per-frame pose information of hands, generated as described in \cite{Zimmermann2017}, and provides this pose information to a Long Short Term Memory network (LSTM) for recognising gestures. However, this method has two  limitations, it can not deal with gestures performed with both hands and all parts of the hand need to be visible in order to get proper pose information, which is not the case with our network.

\section{Ego-Centric Gesture Database}\label{sec:database}

\begin{figure*}[!ht]
\centering
    \includegraphics[scale=0.05]{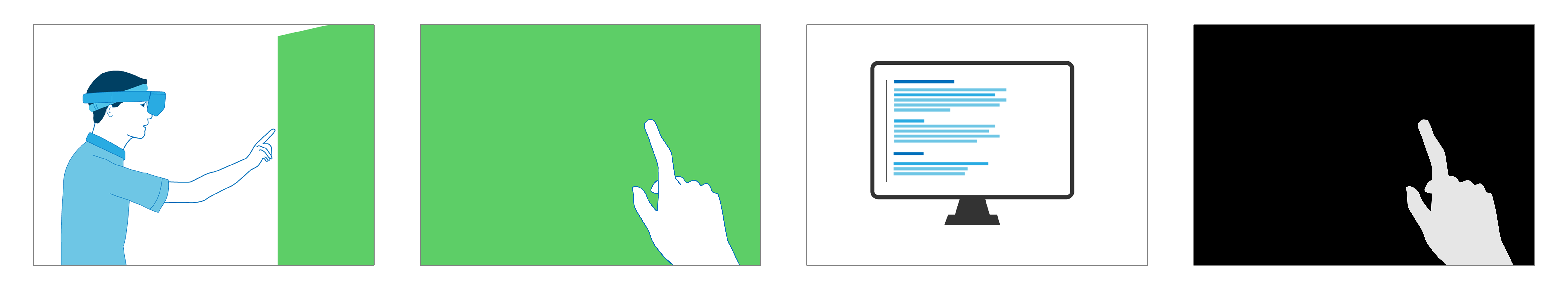}
    \caption{Data Collection Process:: Step 1: User performs gestures in front of a green screen wearing HoloLens. Step 2: Record the images taken from Ego-View camera in HoloLens which sees user's hand performing gestures and a green screen background. Step 3: Transfer frame images to a computer and use a green screen extraction software to generate hand masks for each of the images. Step 4: Save the masks along with their corresponding RGB images. }
    \label{fig:data_collection}
\end{figure*}

\renewcommand{\thesubfigure}{\arabic{subfigure} -}
\addtocounter{subfigure}{-1}

\begin{figure*}[!ht]
 
	\centering 
	\subfigure[Right Shoot]{
        \includegraphics[width=0.39\columnwidth]{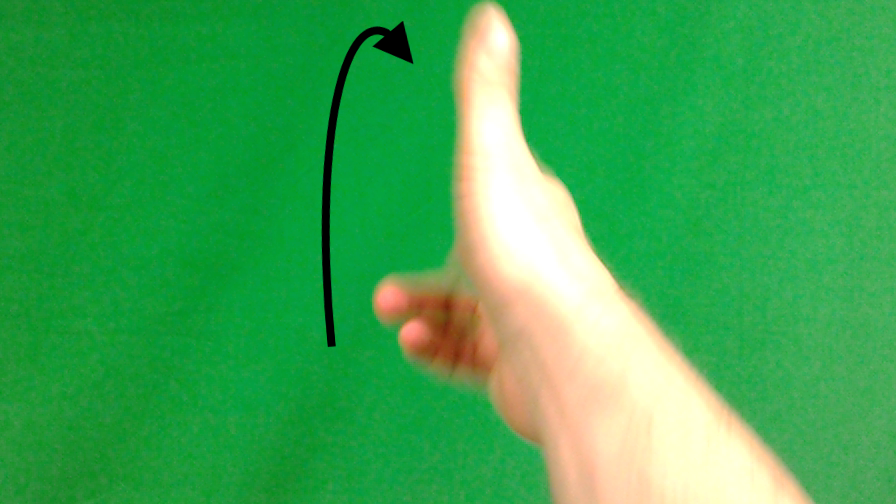}
     }
     \subfigure[Left Shoot]{
        \includegraphics[width=0.39\columnwidth]{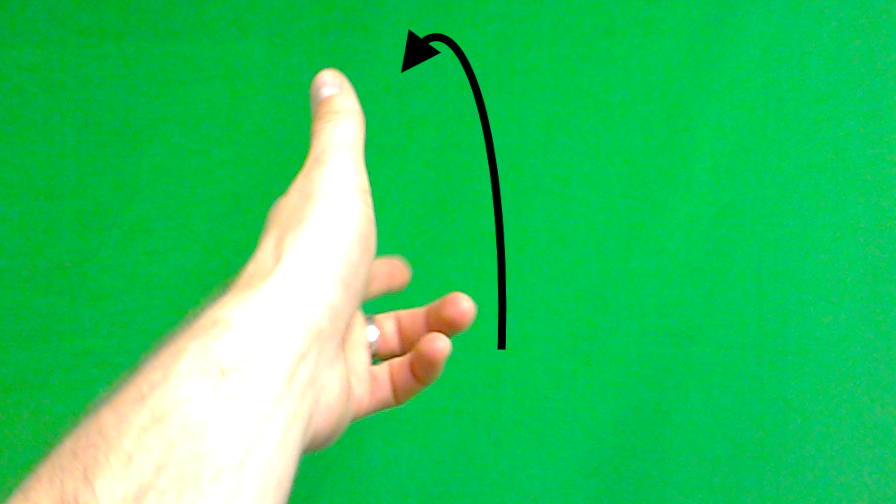}
     }
     \subfigure[Smash]{
        \includegraphics[width=0.39\columnwidth]{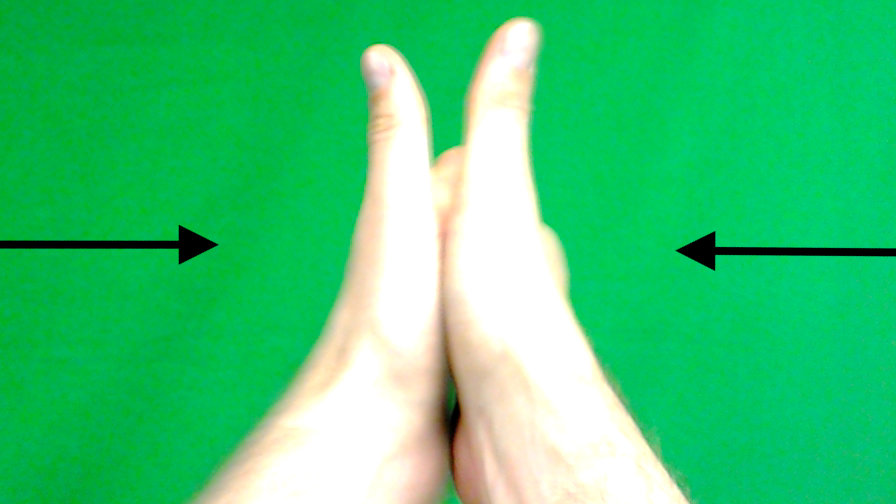}
     }
     \subfigure[Right punch]{
        \includegraphics[width=0.39\columnwidth]{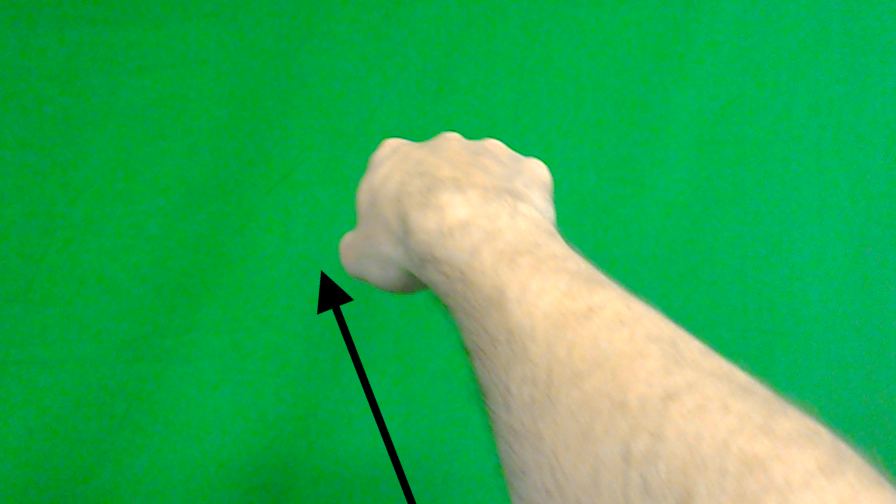}
     }
     \subfigure[Left Punch]{
        \includegraphics[width=0.39\columnwidth]{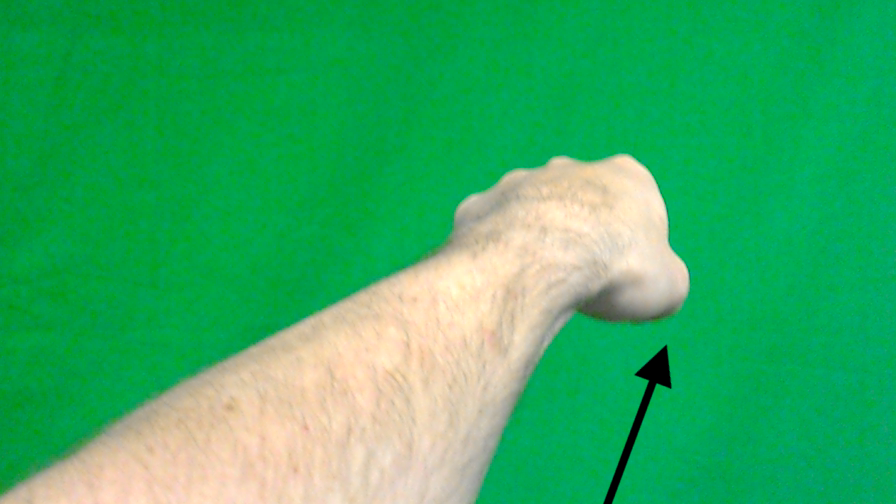}
     }
     \subfigure[Push Back]{
        \includegraphics[width=0.39\columnwidth]{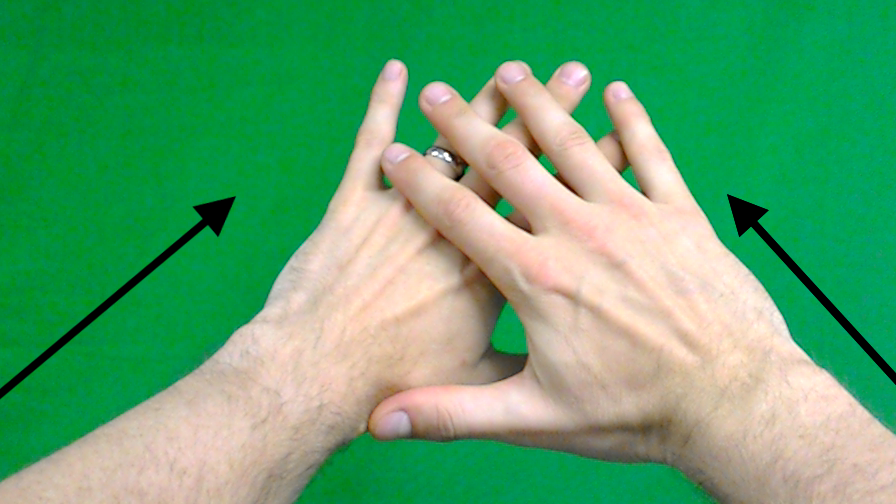}
     }
     \subfigure[Right Block]{
        \includegraphics[width=0.39\columnwidth]{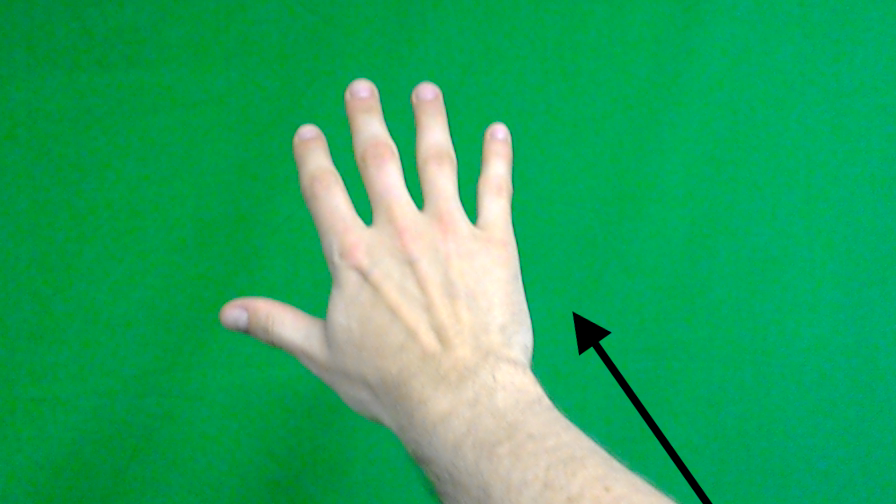}
     }
     \subfigure[Left Block]{
        \includegraphics[width=0.39\columnwidth]{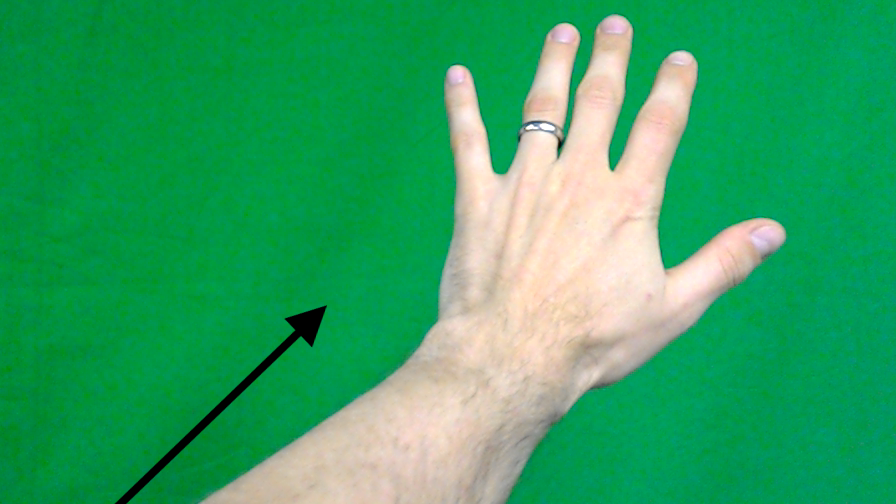}
     }
     \subfigure[Right Teleport]{
        \includegraphics[width=0.39\columnwidth]{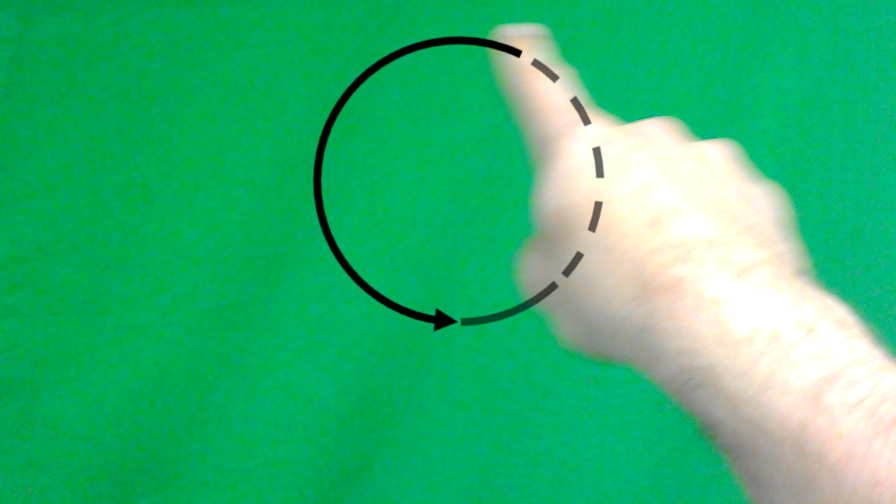}
     }
     \subfigure[Left Teleport]{
        \includegraphics[width=0.39\columnwidth]{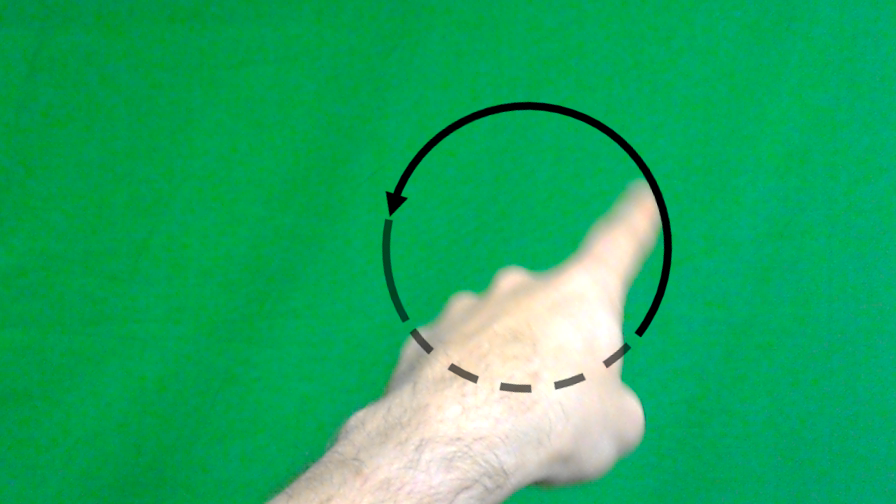}
     }
     
 \caption{Our training set of 10 gestures captured in front of a green screen.}
 \label{fig:gestures}
\end{figure*}

Gestures are usually coupled with a specific task and are rarely the same across different applications. Having a database with a large amount of ego-gestures \cite{Zhang2018} is important to help with the evaluation of different recognition algorithms. However, defining a new gesture is a cumbersome task since it needs to be performed by many subjects in many different scenarios. To alleviate this problem, we used a data augmentation technique which reduces the amount of data that needs to be collected and pre-processed. Table \ref{tab:database_comparison} shows the number of different backgrounds that are needed for our database in comparison to existing ego-hand gesture databases. 

\begin{figure}[tb]
\centering
 \includegraphics[width=0.75\columnwidth]{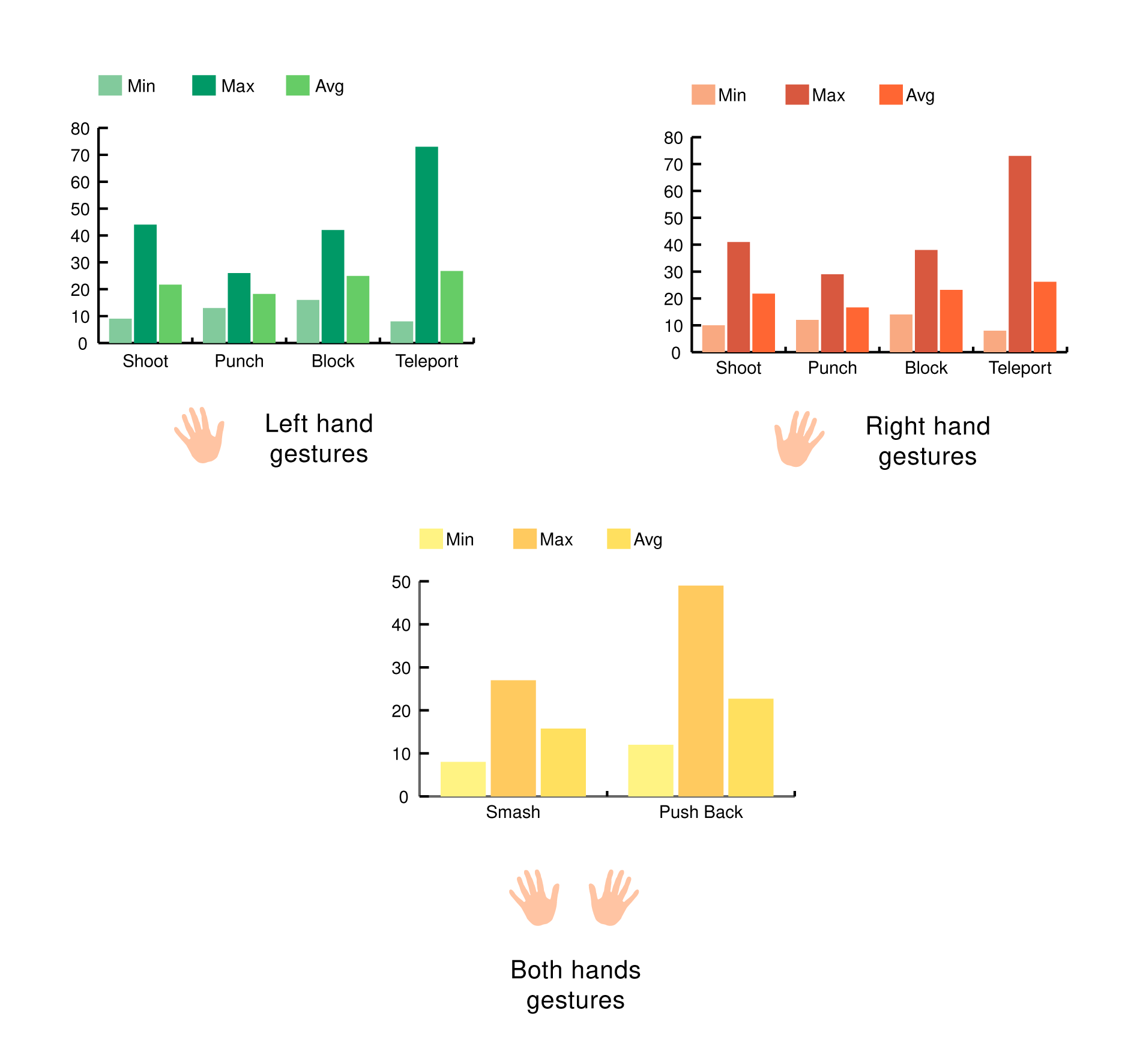}
 \caption{Bar graphs describing the variance in number of frames the user needed to perform a gesture.}
 \label{fig:gesture_variance}
\end{figure}

We defined a set of 10 basic gestures that use left, right and both hands (see Figure~\ref{fig:gestures}). We collected \textbf{training gestures} from 22 users, each repeated the gesture 3 times. Users performed the gesture in front of a green screen wearing HoloLens without any restrictions on the duration of each gesture, allowing them to express naturally. This resulted in a large variance in the duration per gesture and per user, which are described in Figure~\ref{fig:gesture_variance}. In addition to the images (RGB) captured by the egocentric camera, we also collected the 6DOF camera/head pose information that is given by the HoloLens.
\begin{table}[!ht]
\small
\begin{tabular}{|p{2.0cm}||p{1.7cm}|p{2.2cm}| }
\hline
     Database& \# of Subjects & \# of Backgrounds  \\
    \hline
    AirGest\cite{Jain2017} & 6 & 6 \\
    EgoGest\cite{Zhang2018} & 50 & 8 \\
    Ours& 20 & 1 \\
\hline
\end{tabular}
\caption{Table comparing the number of backgrounds needed for existing ego-hand gesture databases with ours. Our database needs only one background compared to others making it much easier to collect data for adding new gestures. }
\label{tab:database_comparison}
\end{table}

The generated images were processed using a green screen segmentation algorithm of a typical video editing software \cite{Natron2018} to generate masks of hands automatically, eliminating the need for manual generation of hand masks per image. Figure~\ref{fig:data_collection} illustrates the process of database generation. The hand masks along with their corresponding images and labels per frame were stored and will be made publicly available.

Unlike the training gestures, we collected \textbf{testing gestures} in natural settings. These gestures were captured in real office environments with various backgrounds. We have collected testing gestures from 6 users, each gesture being repeated twice. After inspecting each video, we removed gestures that were performed outside of camera's field of view and we ended up with 7 to 9 samples per gesture.

To reflect real world situations, our dataset is generated from users with varying skin tones, under different lighting conditions, some users wearing full sleeves, and some wearing watches or bracelets. In comparison, in previously captured datasets (\cite{Zhang2018, Jain2017}), the gestures are more clinical in the sense that each gesture has the same movement of hands or restricted duration. We showed users a video of each gesture at the beginning of the capture and then let them express the gesture naturally.

\section{Network Architecture}\label{sec:architecture}
    The idea of the new architecture is to find spatial feature maps specific to ego-hand gestures and to use these feature maps in a RNN to learn temporal discrimination for recognising ego-hand gestures, while keeping the network small. We achieve this by designing a two stage network architecture.The first stage is Ego-Hand Mask Encoder Network (EHME Net). EHME Net has an hourglass (Fig \ref{fig:architecture}) structure, with a series of convolution filters of increasing depth and decreasing height and width, until they are sufficiently small. Then, we suffix the network with deconvolution filters with decreasing depth and increasing height and width until they reach the size of the mask. We give a detailed description of our EHME Net in Section \ref{ssec:EHME}. Finally, we use the feature maps near the \textit{neck} of the hourglass as an input to an LSTM network to classify the sequence of encoded features. 
    
    The number of parameters that needs to be estimated can approximate the complexity and size of a network. Our network compared to AirGest is much smaller (Table \ref{tab:size_comparision} lists the total number of estimated parameters for our network in comparison to the AirGest network). We believe this could eventually lead to a portable implementation on mobile devices.
    
\begin{table}[!ht]
\small
\begin{tabular}{|p{2.2cm}||p{2.0cm}|}
\hline
     Network& \# of Parameters \\
    \hline
    OurNetwork & 960485\\
    \hline
    AirGest & 17535551 \\
    
\hline
\end{tabular}
\caption{Size of our network in comparison to AirGest \cite{Jain2017} in terms of number of parameters to be estimated.}
\label{tab:size_comparision}
\end{table}
    
    In the subsections \ref{ssec:EHME}, \ref{ssec:sequence_net} we elaborate on the architecture of the module that encodes ego-hand features and the module that recognises gestures from a sequence of encoded ego-hand features respectively.

\subsection{EHME Net} \label{ssec:EHME}
    The first stage of EHME consists of two layers of Resnet~18~\cite{Wu2016}, appended with convolution and pooling layers. The input size of an image is fixed at $224x126$. The series of Resnet, convolution and pooling layers gradually change the feature map size at the \textit{neck} of EHME to $7x4x\textbf{depth}$. The \textbf{depth} of the layer can be varied depending on the gesture dataset. For experiments on our dataset, we fixed the depth to 64. Table~\ref{tab:architecture_parameters} shows all the parameters used to define the shape of EHME used in testing on our gesture dataset. At the end of the neck, we append deconvolution layers to gradually upsample the width and height to size of the mask. And the feature maps' depth decreases to 2 from 64 through the deconvolution layers. At this stage we use a 2D CrossEntropyLoss layer for training, which assigns a probability to each pixel belonging to egocentric hand or not. 
    
    Inspired by \cite{He2017} we add an extension to the \textit{neck} in order to simultaneously generate an ego-hand mask and recognise the encoded features that belong to a particular gesture (see Figure \ref{fig:architecture}). To achieve this, we reduce the size of encoded features to a \textbf{depth} using an average pooling layer and connect this to a fully connected layer of the size of the number of gestures used $N_g$. For training, we also include a 1D CrossEntroyLoss layer at the end. At this point we get a per frame gesture recognition (frame level).

\subsection{Sequence Recognition Net}\label{ssec:sequence_net}
    Frame level recognition can be inaccurate and noisy due to different reasons. Individual images from different gesture sequences can be very similar. In our natural scenario we also have large individual variations for same gestures. Exploiting the temporal dimension and coherence in the data helps to improve the results significantly as we will show in our results (sequence level). Generally, RNNs are known to encode such time-related information well. However, traditional RNNs suffer from the problem of vanishing or exploding gradients. LSTMs are considered an improvement for that problem over traditional RNNs \cite{Hochreiter1997} as they can also forget information that is not relevant over time. Not only do they have better convergence but also provide the ability to be trained on and used with sequences of arbitrary length. This property is crucial for recognition of natural gestures considering the variation in duration needed to express the same gesture by different people (see Figure~\ref{fig:gesture_variance}). Table~\ref{tab:architecture_parameters} shows the parameters of our LSTM that is used for the gesture sequence classification. The hidden layer from the last image in the sequence is connected with a fully connected layer of a size $N_g$ to classify the sequence of gestures.
\section{Training}

We trained our network using the augmented training dataset. The dataset augmentation process is described in  Section~\ref{ssec:data_preprocessing}. The trained network is evaluated using our testing dataset that was collected in natural environments without green screen in the background. This ensures that our network works well on unseen data. The hyper-parameters used for training EHME and Sequence Net are presented in Table~\ref{tab:training_parameters}. In Section~\ref{ssec:training_procedure} we discuss the strategy used for training our network.

\subsection{Data Preprocessing}\label{ssec:data_preprocessing}
\begin{figure}[tb]
 \centering 
 \includegraphics[width=0.9\columnwidth, height=5.0cm]{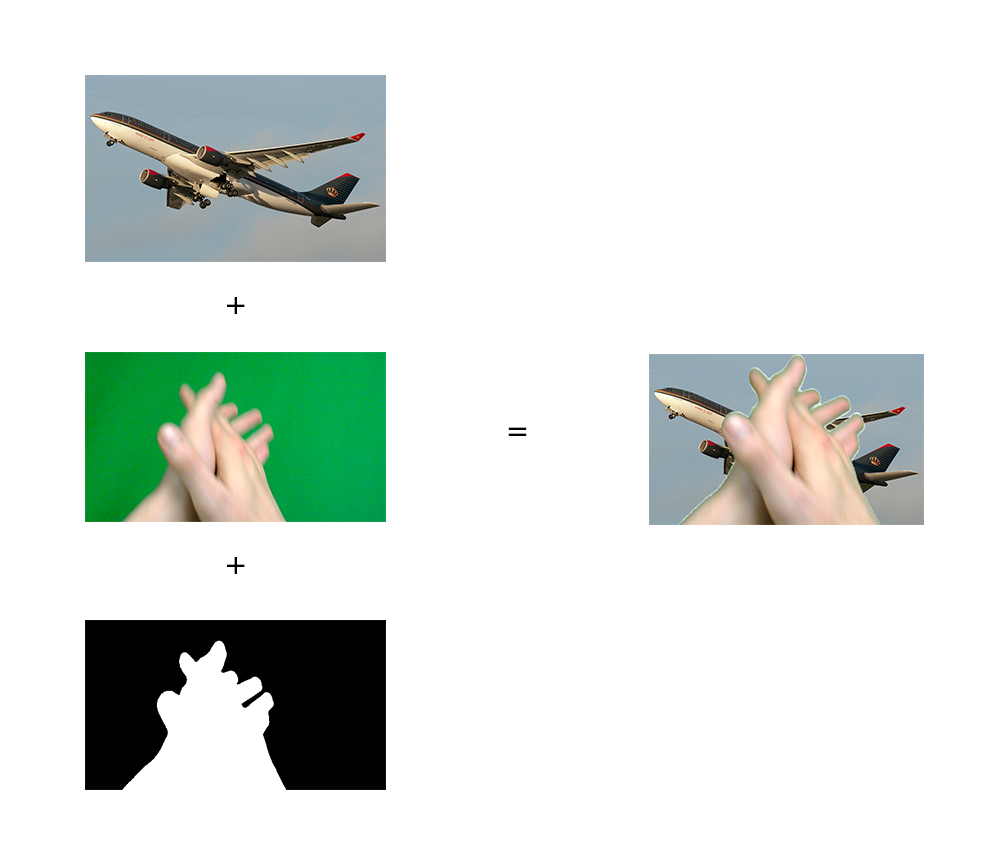}
 \caption{Data augmentation step. The training image on the right is a combination of a random background and a segmented frame using a binary mask (on the left).}
 \label{fig:preprocessing}
\end{figure}

We apply the mask that is obtained by the green screen removal process as described in Section~\ref{sec:database} to the corresponding egocentric image and add $N_i$ backgrounds images to it. Figure~\ref{fig:preprocessing} shows the mask applied to one of the images. This creates $N_i$ images from one captured image with the same mask. As background we chose $N_i$ random images from a set of 40,000 images from the COCO Test dataset \cite{Lin2015}. For our training we set $N_i=5$, which increases the size of dataset fivefold. In addition, we also add one of the following \textit{'none', 'poisson', 'gaussian',} or \textit{'salt\&pepper'} noises randomly, to ensure that we are not over-fitting data. Then, we store each of these images separately along with the gesture id and their corresponding mask. We scale down all the images and masks to \textit{224x126} resolution and normalise them.

\subsection{Training Procedure}\label{ssec:training_procedure}
\begin{figure}[tb]
	\centering 
 \includegraphics[width=0.9\columnwidth]{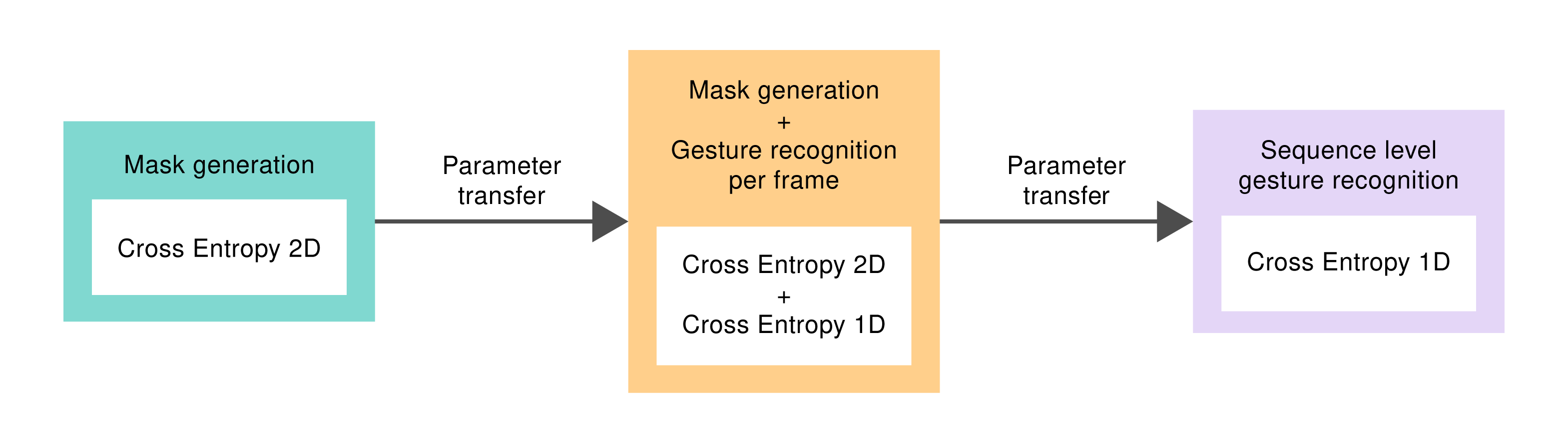}
 \caption{Parameter transfer from various phases of training.}
 \label{fig:training_process}
\end{figure}

The training data is 90/10 split for training and validation respectively throughout the training process. We train the network in 3 phases, the network parameters from each phase are transferred to the next one. This process is described in Figure~\ref{fig:training_process}. In the first phase, we train our EHME defined in Section~\ref{ssec:EHME} with one loss function appended to deconvolution layers learning the ego-hand masks. We use the 2D cross entropy loss and an ADAM optimizer with learning rate $10^{-5}$ for this purpose. The data is shuffled and we train for 5 epochs with batch size of 50. This first phase, in principle trains a hand segmentation network that can actually be used for this purpose as we will show in our experiments with the AirGest dataset described in Section 6.2.

In the second phase, we append an average pooling layer and then a fully connected layer with outputs size $N_g$ to the \textit{neck} of the network. A 1D Cross Entropy Layer is added to do per frame gesture recognition. The parameters obtained from Phase 1 are transferred to Phase 2. Then the network is trained on a combined loss function using 2D cross entropy loss from phase 1 and 1D cross entropy loss from this phase with equal weight for both loss functions. We use ADAM optimiser with learning rate $10^{-6}$ and train for 18 epochs with batch size of 50. At this point we get a frame level gesture recognition framework, which can be inaccurate as explained before and evaluated in our experiments below.

For the final phase, we modify our data augmentation approach. Instead of using a random background and noise for every image, we now use the same augmentation for the whole gesture sequence, such that each gesture sequence has the same background and noise. This is needed for the final phase of training to simulate real conditions. We send a given sequence into the EHME net in a single batch, and save the encoded features as a sequence. Once all the sequences are saved we use these as input to the LSTM. The hidden layer from the last sequence is connected to a fully connected layer with outputs size $N_g$ and then to a 1D cross entropy layer for sequence recognition. All the parameters from earlier phases are used to initialise the weights of EHME + LSTM networks and we do end-to-end training combining the three loss functions. For optimisation, we use a Stochastic Gradient Descent algorithm, with learning rate of $10^{-6}$ and momentum $0.7$. We train for 60 epochs. All the hyper-parameters used for training are summarised in Table~\ref{tab:training_parameters}. After this final phase we get our full network for sequence gesture recognition.

\begin{table}[!ht]
\centering
\tiny
\begin{tabular}{|l||l|l|l|l|} 
\hline
Phase                                                                                 & Optimiser                                                                                              & \begin{tabular}[c]{@{}l@{}}Loss Function\\\end{tabular}                         & \begin{tabular}[c]{@{}l@{}}Batch Size\\\end{tabular} & Epochs  \\ 
\hline
\begin{tabular}[c]{@{}l@{}}Mask Generation\\\end{tabular}                             & \begin{tabular}[c]{@{}l@{}}Adam, $10^{-5}$\\\end{tabular}              & \begin{tabular}[c]{@{}l@{}}2D Cross Entropy\\\end{tabular}                      & 50                                                   & 5       \\ 
\hline
\begin{tabular}[c]{@{}l@{}}Mask Generation\\+\\Frame Level Recognition\\\end{tabular} & \begin{tabular}[c]{@{}l@{}}Adam, lr=$10^{-6}$\\\end{tabular}              & \begin{tabular}[c]{@{}l@{}}2D Cross Entropy\\+\\1D Cross Entropy\\\end{tabular} & 50                                                   & 18      \\ 
\hline
\begin{tabular}[c]{@{}l@{}}Sequence Level Recognition\\\end{tabular}                   & \begin{tabular}[c]{@{}l@{}}SGD, lr=$10^{-6}$, \\momentum=0.7\\\end{tabular} & \begin{tabular}[c]{@{}l@{}}1D Cross Entropy\\\end{tabular}                      & 1                                                    & 60      \\
\hline
\end{tabular}
\caption{Parameters used in various phases of training.}
\label{tab:training_parameters}
\end{table}

\section{Experiments and Results} \label{sec:Experiments}

\begin{figure}[!ht]
	\centering 
	\subfigure[\tiny{Clipped Fingers}]{
        \includegraphics[width=0.25\columnwidth]{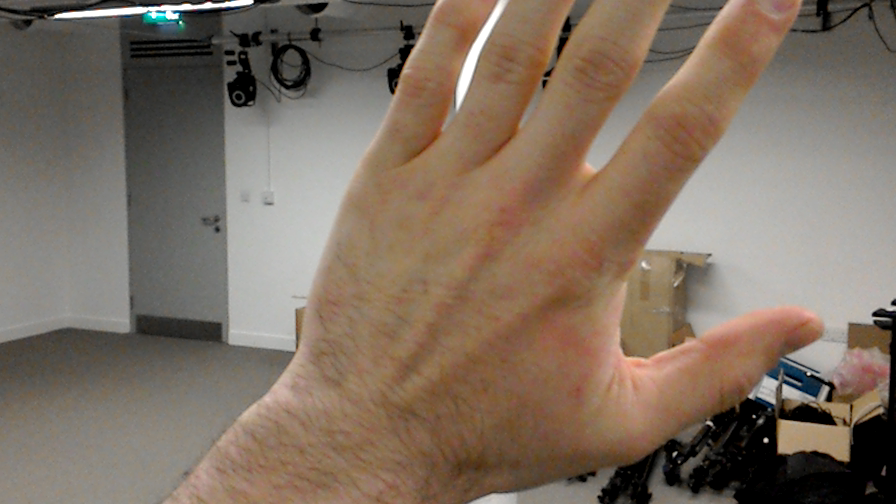}
     }
     \subfigure[\tiny{Strong Motion Blur}]{
        \includegraphics[width=0.25\columnwidth]{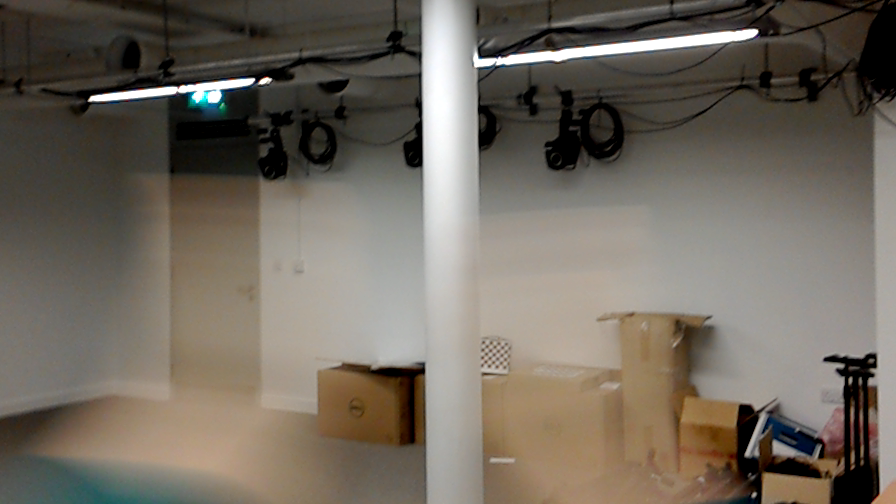}
     }
      \subfigure[\tiny{Gesture Variation}]{
        \includegraphics[width=0.25\columnwidth]{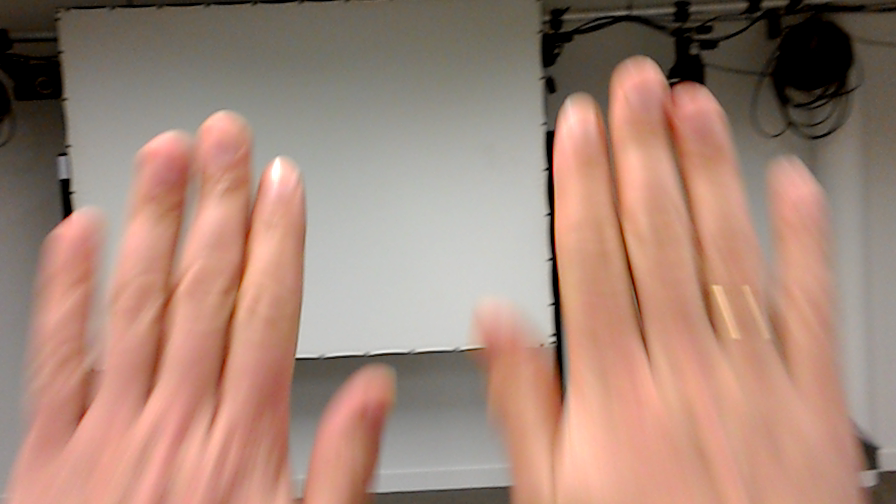}
     }
 \caption{Examples of frames from testing sequences. \textbf{a)} Left Block gesture that was not performed fully inside the camera FOV. \textbf{b)} Smash gesture done at high speed with a strong motion blur effect. \textbf{c)} Push Back gesture that is a variation to the one in the training dataset. }
 \label{fig:natural_example}
\end{figure}

\begin{table}[!ht]
\small
\begin{tabular}{|p{2.0cm}||p{1.3cm}|p{1.3cm}|p{1.0cm}|}
\hline
     Network& \# of Gestures Classified Correctly & \# of Gestures Classified Wrongly & Accuracy \%  \\
    \hline
    Frame Level& 49 & 35 & 58.33 \\
    Sequence Level & 60 & 24 & 71.42 \\
\hline
\end{tabular}
\caption{Accuracy results for gesture recognition on our dataset}
\label{tab:results_our_dataset}
\end{table}
\begin{table*}
\tiny
\centering

\begin{tabular}{|l|l|l|llllll}
\hline
\multicolumn{9}{|c|}{Network Parameters}                                                                        \\ \hline
\multicolumn{3}{|c|}{\cellcolor[HTML]{9FE5F7}EHME}                                              & \multicolumn{3}{c|}{\cellcolor[HTML]{FFCF8B}Encoded Features}                                                                                                                          & \multicolumn{3}{c|}{\cellcolor[HTML]{E4D6F7}Sequence Net}                                                                                                                                             \\ \hline
\cellcolor[HTML]{9FE5F7}Node Type                    & \cellcolor[HTML]{9FE5F7}Output Size                 & \cellcolor[HTML]{9FE5F7}Node Parameters                                                        & \multicolumn{1}{l|}{\cellcolor[HTML]{FFCF8B}Node Type}       & \multicolumn{1}{l|}{\cellcolor[HTML]{FFCF8B}Output Size} & \multicolumn{1}{l|}{\cellcolor[HTML]{FFCF8B}Node Parameters} & \multicolumn{1}{l|}{\cellcolor[HTML]{E4D6F7}Node Type}       & \multicolumn{1}{l|}{\cellcolor[HTML]{E4D6F7}Output Size} & \multicolumn{1}{l|}{\cellcolor[HTML]{E4D6F7}Node Parameters}                \\ \hline
\cellcolor[HTML]{9FE5F7}                             & \cellcolor[HTML]{9FE5F7}                            & \multicolumn{1}{c|}{\cellcolor[HTML]{9FE5F7}}                                                  &                                                              &                                                          &                                                              &                                                              &                                                          &                                                                             \\
\multirow{-2}{*}{\cellcolor[HTML]{9FE5F7}Resnet18 1} & \multirow{-2}{*}{\cellcolor[HTML]{9FE5F7}112x63x64} & \multicolumn{1}{c|}{\cellcolor[HTML]{9FE5F7}}                                                  &                                                              &                                                          &                                                              &                                                              &                                                          &                                                                             \\ \cline{1-2}
\cellcolor[HTML]{9FE5F7}Resnet18 2                   & \cellcolor[HTML]{9FE5F7}56x32x64                    & \multicolumn{1}{c|}{\cellcolor[HTML]{9FE5F7}}                                                  &                                                              &                                                          &                                                              &                                                              &                                                          &                                                                             \\ \cline{1-2}
\cellcolor[HTML]{9FE5F7}Resnet18 3                   & \cellcolor[HTML]{9FE5F7}28x16x128                   & \multicolumn{1}{c|}{\multirow{-4}{*}{\cellcolor[HTML]{9FE5F7}Parameters from \cite{Wu2016}}} &                                                              &                                                          &                                                              &                                                              &                                                          &                                                                             \\ \cline{1-3}
\cellcolor[HTML]{9FE5F7}conv1                        & \cellcolor[HTML]{9FE5F7}14x8x128                    & \cellcolor[HTML]{9FE5F7}3x3, stride 2, padding 1                                               &                                                              &                                                          &                                                              &                                                              &                                                          &                                                                             \\ \hline
\cellcolor[HTML]{9FE5F7}conv2                        & \cellcolor[HTML]{9FE5F7}7x4x64                      & \cellcolor[HTML]{9FE5F7}3x3, stride 2, padding 1                                               & \multicolumn{1}{l|}{\cellcolor[HTML]{FFCF8B}Average Pool}    & \multicolumn{1}{l|}{\cellcolor[HTML]{FFCF8B}1x1x64}      & \multicolumn{1}{l|}{\cellcolor[HTML]{FFCF8B}7x4}             & \multicolumn{1}{l|}{\cellcolor[HTML]{E4D6F7}LSTM}            & \multicolumn{1}{l|}{\cellcolor[HTML]{E4D6F7}64}          & \multicolumn{1}{l|}{\cellcolor[HTML]{E4D6F7}input 64, hidden 128, layers 3} \\ \hline
\cellcolor[HTML]{80D8D0}deconv1                      & \cellcolor[HTML]{80D8D0}14x8x32                     & \cellcolor[HTML]{80D8D0}4x4, stride 2, padding 1                                               & \multicolumn{1}{l|}{\cellcolor[HTML]{FFCF8B}Fully Connected} & \multicolumn{1}{l|}{\cellcolor[HTML]{FFCF8B}1x11}        & \multicolumn{1}{l|}{\cellcolor[HTML]{FFCF8B}64x11}           & \multicolumn{1}{l|}{\cellcolor[HTML]{E4D6F7}Fully Connected} & \multicolumn{1}{l|}{\cellcolor[HTML]{E4D6F7}10}          & \multicolumn{1}{l|}{\cellcolor[HTML]{E4D6F7}64x10}                          \\ \hline
\cellcolor[HTML]{80D8D0}deconv2                      & \cellcolor[HTML]{80D8D0}28x16x16                    & \cellcolor[HTML]{80D8D0}4x4, stride 2, padding 1                                               &                                                              &                                                          &                                                              &                                                              &                                                          &                                                                             \\ \cline{1-3}
\cellcolor[HTML]{80D8D0}deconv3                      & \cellcolor[HTML]{80D8D0}56x32x8                     & \cellcolor[HTML]{80D8D0}4x4, stride 2, padding 1                                               &                                                              &                                                          &                                                              &                                                              &                                                          &                                                                             \\ \cline{1-3}
\cellcolor[HTML]{80D8D0}deconv4                      & \cellcolor[HTML]{80D8D0}112x64x4                    & \cellcolor[HTML]{80D8D0}4x4, stride 2, padding 1                                               &                                                              &                                                          &                                                              &                                                              &                                                          &                                                                             \\ \cline{1-3}
\cellcolor[HTML]{80D8D0}deconv5                      & \cellcolor[HTML]{80D8D0}224x126x2                   & \cellcolor[HTML]{80D8D0}4x4, stride 2, padding (2,1)                                           &                                                              &                                                          &                                                              &                                                              &                                                          &                                                                             \\ \cline{1-3}
\end{tabular}
\caption{All the parameters used for various modules of the network architecture.}
\label{tab:architecture_parameters}
\end{table*}
We have tested our network architecture on our \textit{testing database} and on a dataset from AirGest~\cite{Jain2017}. Our testing dataset contains 10 natural gestures, where fingers are clipped, frames have a strong motion blur, and there is a variation within a gesture (see Figure \ref{fig:natural_example} for examples). The AirGest dataset contains 4 gestures (click, bloom, zoom out and zoom in) that are performed in a clinical manner, i.e gestures have clear separation between phases, they are performed slowly without motion blur, and are fully contained in the center of the video.

In the following section we present and discuss the results from various phases of our network on our dataset, and in Section~\ref{ssec:results_airgest} we present comparative results of our network on the AirGest dataset.
Our network architecture is implemented in PyTorch and we used a PC with an Intel Core i7 CPU and NVidia Titan Xp GPU for both, training and testing.

\renewcommand{\thesubfigure}{(\alph{subfigure})}

\subsection{Recognition on our dataset}\label{ssec:results_our_dataset}

As mentioned in Section~\ref{ssec:training_procedure} we used 3 phases for training. In our testing we report results of networks from Phase 2 and Phase 3. The network result obtained from Phase 2 of training can perform recognition per frame. The input to this network is a sequence of RGB images of a gesture, and we get a gesture classification for each frame. A simple voting strategy is followed giving each gesture a vote if a frame is predicted to be that gesture. The sequence is then assigned the gesture with maximum number of votes. We call this \textit{frame level} recognition.

For \textit{sequence level} recognition we input a sequence of images to the network from Phase 3. As we can observe from the results in Table~\ref{tab:results_our_dataset} sequence level recognition performs much better than frame level recognition. The same hand pose can be part of multiple different gestures, but during different stages of performing the gesture. Since frame level recognition does not consider time, it could easily misclassify a gesture. Adding a temporal recognition component like an LSTM solves this issue as is evident from the results in Table~\ref{tab:results_our_dataset}.

\begin{figure}[!htb]
 \centering 
 \includegraphics[width=0.7\columnwidth]{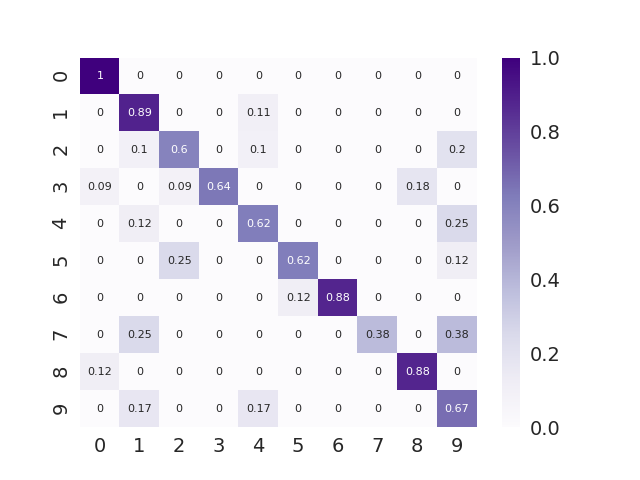}
 \caption{Normalised Confusion Matrix for 10 gestures in our database. X-axis has the predicted labels and Y-axis the ground truth labels}
 \label{fig:confusion_matrix}
\end{figure}

To analyse recognition performance on each gesture we present a normalised confusion matrix in Figure~\ref{fig:confusion_matrix} for results from the \textit{sequence level} recognition. The mislabelled gestures are within the same hand (as in a left-handed gesture is being labelled as another left-handed gesture but not a right-handed). The recognition of gesture 7 - Left Block is especially low and is confused with Left Shoot and Teleport gestures. Looking at the testing videos closely, one observation that could explain this confusion is a large head movement that creates relative motion inside the frame similar to the one in Shoot (up-left motion) and Teleport (circular motion in left direction). To improve the accuracy in these situations we are planing in the future to utilise the head pose transformation that can be obtained from HoloLens.

Also, in the Teleport gesture, some users used the whole arm to create circular motion, where others used only one finger. This small finger-motion is especially challenging to distinguish from an egocentric view as it can be occluded by the arm or hand, and can be easily confused with Shoot or Punch gestures. This is something that should be taken into consideration when designing gestures for egocentric view recognition.

The network to find hand pose \cite{Zimmermann2017} which was used in AirGestAR \cite{Jain2017} could not recover poses for many of the ego-hand images in our dataset, as it was not designed to handle complex scenarios like motion blur and clipped fingers which frequently occur in our dataset. Our network however could handle such situations which was not the case with AirGestAR network strategy. So we could not perform a comparative study using their network.

\subsection{Recognition on AirGest Dataset}\label{ssec:results_airgest}
To compare our network with previous work we used a dataset from AirGest \cite{Jain2017}. This dataset does not have hand masks as ground truth, which are needed to perform Phase 1 training in our network. To avoid manual mask extraction, we used our Phase 1 network that is trained with our training dataset to generate these masks automatically. After visual verification, we used these masks as a ground truth in addition to frame level labels in Phase 2 training. Finally, we followed the same procedure mentioned in section \ref{ssec:training_procedure} for Phase 3 training. 

\begin{figure}[!ht]
	\centering 
	\subfigure[Ours with $\sigma= 0.5$]{
        \includegraphics[width=0.35\columnwidth]{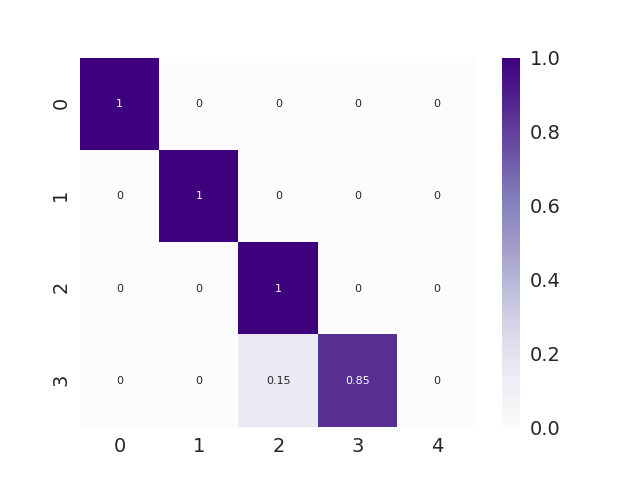}
     }
     \subfigure[AirGest $\sigma = 0.7$]{
        \includegraphics[width=0.4\columnwidth]{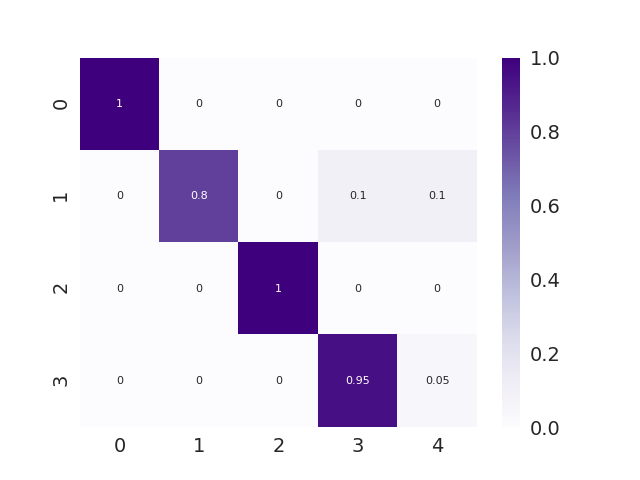}
     }
 \caption{Confusion matrix for AirGest dataset.}
 \label{fig:airgest_conf_mat}
\end{figure}

To provide comparative results we used the same training and testing data as described in \cite{Jain2017}. The confusion matrix is presented in Figure~\ref{fig:airgest_conf_mat} and the overall accuracy in Table~\ref{tab:results_airgest}. Our final network's performance is able to match the AirGest network's despite being smaller in size.

\begin{table}[!ht]
\small
\centering
\begin{tabular}{|p{0.8cm}||p{1.3cm}|p{1.3cm}|p{1.0cm}|p{1.0cm}|p{0.8cm}| }
\hline
     Network& \# of Gestures Classified Correctly & \# of Gestures Classified Wrongly & Unclassified & Probability Threshold ($\sigma$)& Accuracy \%  \\
    \hline
    Ours & 77 & 3 & 0 & 0.5 & 96.25 \\
    Ours & 75 & 3 & 2 & 0.66 & 93.75 \\
    \hline
    AirGest& 75 & 2 & 3 & 0.7 & 93.75 \\
\hline
\end{tabular}
\caption{Accuracy results for our network and network from \cite{Jain2017} on AirGest dataset. Following the procedure used in \cite{Jain2017} we experimented with different probability thresholds to mark a gesture sequence as classified vs unclassified. Any sequence with probability lower than the mentioned threshold is marked as unclassified.}
\label{tab:results_airgest}
\end{table}

\renewcommand{\thesubfigure}{}
\section{Conclusion \& Future Work}
We propose a novel deep learning network architecture which simultaneously encodes ego-hands in a sequence of images and recognises the egocentric gesture. A novel data augmentation technique using green screening to decrease the burden of collecting large amounts of data from a large number of users is introduced. The network architecture in conjunction with the data augmentation technique makes adding a new egocentric gesture for recognition easier. In addition, we also publish our training and testing dataset with 10 gestures performed in a less clinical and a less constrained manner. We evaluate our network which is trained on the augmented dataset and tested on a natural (i.e. gestures performed not in front of green screen) dataset. Our network can deal with a variations in the gestures' length, style and motion blur as presented in the results.

Recognising gestures on continuous video is also essential for making natural interactions possible on head-mounted AR devices. Handling this challenging task is one of the directions we want to explore in the future. Another direction we want to seek is to use head pose information and other modalities provided by the AR devices to deal with sudden and extreme head motion, paving the way for recognition of more complicated gestures in difficult scenarios and natural activities (e.g. GTEA Gaze+ \cite{Li2015}).

\acknowledgments{
This publication has emanated from research conducted with the financial support of Science Foundation Ireland (SFI) under the Grant Number 15/RP/2776. Sincere thanks to Sahitya Parvathaneni for doing the major part of illustrations.}

\bibliographystyle{abbrv-doi}


\begin{thebibliography}{10}

\bibitem{Baraldi2014}
L.~Baraldi, F.~Paci, G.~Serra, L.~Benini, and R.~Cucchiara.
\newblock {Gesture recognition in ego-centric videos using dense trajectories
  and hand segmentation}.
\newblock {\em IEEE Computer Society Conference on Computer Vision and Pattern
  Recognition Workshops}, pp. 702--707, 2014.

\bibitem{Cao2017}
C.~Cao, Y.~Zhang, Y.~Wu, H.~Lu, and J.~Cheng.
\newblock {Egocentric Gesture Recognition Using Recurrent 3D Convolutional
  Neural Networks with Spatiotemporal Transformer Modules}.
\newblock {\em 2017 IEEE International Conference on Computer Vision (ICCV)},
  pp. 3783--3791, 2017.

\bibitem{Cutler1998}
R.~Cutler and M.~Turk.
\newblock {View-based interpretation of real-time optical flow for gesture
  recognition}.
\newblock {\em Proceedings of IEEE International Conference on Automatic Face
  and Gesture Recognition}, pp. 416--421, 1998.

\bibitem{Donahue2014}
J.~Donahue, L.~A. Hendricks, S.~Guadarrama, M.~Rohrbach, S.~Venugopalan,
  K.~Saenko, and T.~Darrell.
\newblock {Long-term recurrent convolutional networks for visual recognition
  and description}.
\newblock {\em IEEE Conference on Computer Vision and Pattern Recognition
  (CVPR)}, 39(4):677--691, 2015.

\bibitem{Girshick2015}
R.~Girshick.
\newblock {Fast R-CNN}.
\newblock {\em Proceedings of the IEEE International Conference on Computer
  Vision}, 2015 Inter:1440--1448, 2015.

\bibitem{Goodale1992}
M.~a. Goodale and a.~D. Milner.
\newblock {Separate visual pathways for perception and action.}
\newblock {\em Trends in Neurosciences}, 15(I):20--5, 1992.

\bibitem{He2017}
K.~He, G.~Gkioxari, P.~Doll{\'{a}}r, and R.~Girshick.
\newblock {Mask R-CNN}.
\newblock {\em Proceedings of the IEEE International Conference on Computer
  Vision}, 2017.

\bibitem{david2016}
D.~Held, S.~Thrun, and S.~Savarese.
\newblock {GoTurn:Learning to Track at 100 FPS with Deep Regression Networks}.
\newblock {\em European Conference on Computer Vision (ECCV)}, 2016.

\bibitem{Hochreiter1997}
S.~Hochreiter and J.~{Urgen Schmidhuber}.
\newblock {Long Short-Term Memory}.
\newblock {\em Neural Computation}, 9(8):1735--1780, 1997.

\bibitem{Jain2017}
V.~Jain, R.~Perla, and R.~Hebbalaguppe.
\newblock {AirGestAR: Leveraging Deep Learning for Complex Hand Gestural
  Interaction with Frugal AR Devices}.
\newblock {\em Adjunct Proceedings of the 2017 IEEE International Symposium on
  Mixed and Augmented Reality, ISMAR-Adjunct 2017}, pp. 235--239, 2017.

\bibitem{Krizhevsky2012}
A.~Krizhevsky, I.~Sutskever, and G.~E. Hinton.
\newblock {ImageNet Classification with Deep Convolutional Neural Networks}.
\newblock {\em Advances In Neural Information Processing Systems}, pp. 1--9,
  2012.

\bibitem{Li2015}
Y.~Li, Z.~Ye, and J.~M. Rehg.
\newblock {Delving into egocentric actions}.
\newblock {\em Proceedings of the IEEE Computer Society Conference on Computer
  Vision and Pattern Recognition}, pp. 287--295, 2015.

\bibitem{Lin2015}
T.-Y. Lin, C.~L. Zitnick, and P.~Doll.
\newblock {Microsoft COCO : Common Objects in Context}.
\newblock {\em Arixiv}, pp. 1--15, 2015.

\bibitem{Molchanov2016}
P.~Molchanov, X.~Yang, S.~Gupta, K.~Kim, S.~Tyree, and J.~Kautz.
\newblock {Online Detection and Classification of Dynamic Hand Gestures with
  Recurrent 3D Convolutional Neural Networks}.
\newblock {\em IEEE Conference on Computer Vision and Pattern Recognition
  (CVPR)}, pp. 4207--4215, 2016.

\bibitem{Moni2009}
M.~A. Moni and A.~B.~M. {Shawkat Ali}.
\newblock {HMM based hand gesture recognition: A review on techniques and
  approaches}.
\newblock {\em Proceedings - 2009 2nd IEEE International Conference on Computer
  Science and Information Technology}, pp. 433--437, 2009.

\bibitem{Mu-ChunSu2000}
{Mu-Chun Su}.
\newblock {A fuzzy rule-based approach to spatio-temporal hand gesture
  recognition}.
\newblock {\em IEEE Transactions on Systems, Man and Cybernetics, Part C
  (Applications and Reviews)}, 30(2):276--281, 2000.

\bibitem{Natron2018}
Natron.
\newblock Natron.
\newblock {\em www.natron.fr}, 2018.

\bibitem{Posner2016}
I.~Posner and P.~Ondruska.
\newblock {Deep Tracking: Seeing Beyond Seeing Using Recurrent Neural
  Networks}.
\newblock {\em Proceedings of the 30th Conference on Artificial Intelligence
  (AAAI 2016)}, pp. 3361--3367, 2016.

\bibitem{Simonyan2014}
K.~Simonyan and A.~Zisserman.
\newblock {Two-Stream Convolutional Networks for Action Recognition in Videos}.
\newblock {\em Advances in Neural Information Processing Systems}, pp. 1--11,
  2014.

\bibitem{Simonyan2014a}
K.~Simonyan and A.~Zisserman.
\newblock {Very Deep Convolutional Networks for Large-Scale Image Recognition}.
\newblock {\em CoRR}, abs/1409.1:1--14, 2014.

\bibitem{Singh2016}
S.~Singh, C.~Arora, and C.~V. Jawahar.
\newblock {First Person Action Recognition Using Deep Learned Descriptors}.
\newblock {\em IEEE Conference on Computer Vision and Pattern Recognition
  (CVPR)}, pp. 2620--2628, 2016.

\bibitem{Tran2015}
D.~Tran, L.~Bourdev, R.~Fergus, L.~Torresani, and M.~Paluri.
\newblock {Learning spatiotemporal features with 3D convolutional networks}.
\newblock {\em Proceedings of the IEEE International Conference on Computer
  Vision}, 2015 Inter:4489--4497, 2015.

\bibitem{Wang2013}
H.~Wang, C.~Schmid, A.~Recognition, and T.~Iccv.
\newblock {Action Recognition with Improved Trajectories}.
\newblock {\em Proceedings of the IEEE International Conference on Computer
  Vision}, pp. 3551--3558, 2013.

\bibitem{Wang2017b}
P.~Wang, W.~Li, S.~Liu, Z.~Gao, C.~Tang, and P.~Ogunbona.
\newblock {Large-scale Isolated Gesture Recognition Using Convolutional Neural
  Networks}.
\newblock {\em IEEE International Conference on Pattern Recognition}, pp.
  19--24, 2017.

\bibitem{Wu2016}
S.~Wu, S.~Zhong, and Y.~Liu.
\newblock {Deep Residual Learning for Image Recognition}.
\newblock {\em IEEE Conference on Computer Vision and Pattern Recognition
  (CVPR)}, pp. 1--17, 2016.

\bibitem{Zhang2018}
Y.~Zhang, C.~Cao, J.~Cheng, and H.~Lu.
\newblock {EgoGesture: A New Dataset and Benchmark for Egocentric Hand Gesture
  Recognition}.
\newblock {\em IEEE Transactions on Multimedia}, 9210(c):1--1, 2018.

\bibitem{Zimmermann2017}
C.~Zimmermann and T.~Brox.
\newblock {Learning to Estimate 3D Hand Pose from Single RGB Images}.
\newblock {\em Proceedings of the IEEE International Conference on Computer
  Vision}, 2017.

\end{thebibliography}

\end{document}